\documentclass{isprs} 
\usepackage{subfigure}
\usepackage{setspace}
\usepackage{geometry} 
\usepackage{epstopdf}
\usepackage[labelsep=period]{caption}  
\usepackage[british]{babel} 
\usepackage[hang]{footmisc}

\usepackage{siunitx}    
\usepackage{rotating}
\usepackage{array,multirow}
\usepackage{adjustbox}
\usepackage{colortbl}
\usepackage{xcolor}
\usepackage{arydshln}
\usepackage{caption}
\usepackage{todonotes}
\usepackage{natbib}

\begin{document}

\title{Segmentation of Industrial Burner Flames: A Comparative Study from Traditional Image Processing to Machine and Deep Learning}

\author{S. Landgraf\textsuperscript{1,}\footnotemark \hspace{1 mm}, M. Hillemann\textsuperscript{1}, M. Aberle\textsuperscript{2}, V. Jung\textsuperscript{2}, M. Ulrich\textsuperscript{1}}

\address{\textsuperscript{1, 2}Institute of Photogrammetry and Remote Sensing (IPF), Karlsruhe Institute of Technology (KIT), Germany -\\ \textsuperscript{1 }(steven.landgraf, markus.hillemann, markus.ulrich)@kit.edu \\
\textsuperscript{2 }(moritz.aberle, valentin.jung)@student.kit.edu \\}


\abstract{
In many industrial processes, such as power generation, chemical production, and waste management, accurately monitoring industrial burner flame characteristics is crucial for safe and efficient operation. A key step involves separating the flames from the background through binary segmentation. Decades of machine vision research have produced a wide range of possible solutions, from traditional image processing to traditional machine learning and modern deep learning methods. In this work, we present a comparative study of multiple segmentation approaches, namely Global Thresholding, Region Growing, Support Vector Machines, Random Forest, Multilayer Perceptron, U-Net, and DeepLabV3+, that are evaluated on a public benchmark dataset of industrial burner flames. We provide helpful insights and guidance for researchers and practitioners aiming to select an appropriate approach for the binary segmentation of industrial burner flames and beyond. For the highest accuracy, deep learning is the leading approach, while for fast and simple solutions, traditional image processing techniques remain a viable option.\footnotemark
}

\maketitle

\section{Introduction}\label{Introduction}
\footnotetext{\textsuperscript{*}Corresponding author}
\footnotetext{\textsuperscript{†}Our modified version of the used dataset:\\https://publikationen.bibliothek.kit.edu/1000159497}
Industrial combustion processes play a critical role in many industrial processes, including power generation, chemical production, and waste management. To ensure safe and efficient operation, it is essential to accurately monitor the characteristics of the burner flames, such as their shape, size, and temperature. A key step in camera-based monitoring involves separating the flames from the background through segmentation \citep{grosskopf2021evaluation,landgraf2022evaluation}.

Traditionally, binary segmentation has been performed using image processing techniques such as thresholding, edge detection, region growing, and morphological operations \citep{steger2018MachineVision}. While these techniques can be effective, they often require deep knowledge about the application, significant manual tuning, and may not generalize well to new data.

In recent years, deep learning has emerged as a powerful data-driven approach for segmentation tasks. These methods can learn meaningful features and patterns from training data, which generalize well to the real application, and therefore achieve higher accuracy than traditional techniques \citep{long2015fully}. However, they require large amounts of training data, which are costly to label, and are typically computationally more expensive. 

In this work, we present a comparative study of traditional image processing techniques, traditional machine learning-based methods, and deep learning approaches for binary segmentation of industrial burner flames in grayscale images. 
Our study provides insights into the strengths and limitations of each approach and can help guide researchers and practitioners in selecting the most appropriate method for their specific application.


\begin{figure}[t!]
    \centering
    \includegraphics[width=0.95\linewidth]{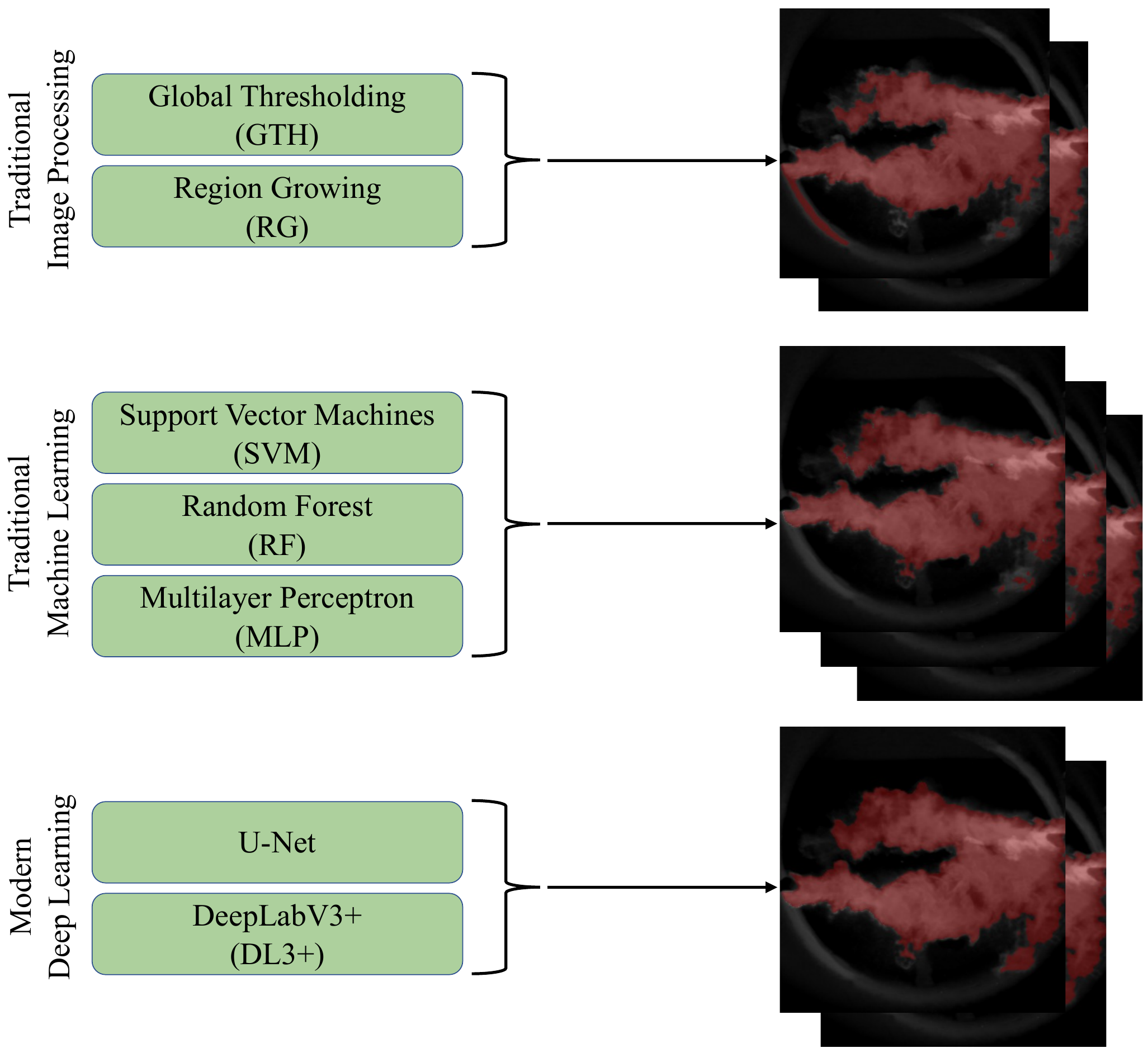}
    \caption{An overview of our comparative study. We compare seven different approaches for the binary segmentation of industrial burner flames: Global Thresholding, Region Growing, Support Vector Machines, Random Forest, Multilayer Perceptron, U-Net, and DeepLabV3+.}
    \label{fig:overview}
\end{figure}


\section{Related Work}\label{Related Work}

The following related work focuses on flame segmentation and is divided into three categories: i) Traditional image processing methods, ii) traditional machine learning methods, and iii) deep learning methods. With traditional image processing methods, we refer to methods that use hand-crafted features and rule-based decisions to distinguish between flame and background. Traditional machine learning methods, on the other hand, refer to data-driven methods like Support Vector Machines (SVM), Random Forest (RF), and Multilayer Perceptron (MLP) which work with hand-crafted features. Finally, by deep learning methods, we mean state-of-the art methods for image segmentation with modern deep neural network-based architectures. This section only covers related work regarding the segmentation of flames in images. Classification and detection approaches are not covered to limit the scope.

\textbf{Traditional Image Processing.}
Early approaches for flame segmentation use traditional image processing techniques \citep{celik2007fire,zhang2009color,fan2014flame,wang2014flame,matthes2019new,li2020flame}. These methods rely on different hand-crafted features such as color \citep{celik2007fire,zhang2009color,wang2014flame,li2020flame} and geometrical characteristics like area, roundness and contour fluctuation \citep{zhang2009color}. Several of these methods use image sequences or videos of the flames to determine time-dependent features like color and area changes \citep{zhang2009color,wang2014flame,matthes2019new,li2020flame}. Some methods distinguish between flame and background by using empirically determined thresholds for combinations of these features \citep{zhang2009color,wang2014flame} or determine the thresholds automatically using Otsu's method \citep{matthes2019new}. Others combine the features with fuzzy logic \citep{celik2007fire} or a Bayesian model \citep{li2020flame}. \citet{fan2014flame} choose a different approach and use a level set method to detect the contour of the flame.

\textbf{Traditional Machine Learning.}
Many of the traditional machine learning methods for flame segmentation also use hand-crafted features like those described above. However, these methods do not decide based on rules created by the human developer, but create their own set of rules based on the data. In addition to some of the features described above, \citet{borges2010probabilistic} use surface coarseness, boundary roughness and skewness as geometrical characteristics and a Bayesian Classifier flame segmentation. \citet{zhao2011svm} and \citet{jamali2013outdoor} also use image sequences and time-dependent features and then perform segmentation with a SVM, where \citet{zhao2011svm} takes into account other texture features such as entropy and contrast. Recently, \citet{liang2022random} proposed a two step segmentation method. In the first step, they determine the feature importance of different geometric, texture and color features with a RF, and in the second step they use the five most important features as input neurons of a MLP that distinguishes between flame and background.

\textbf{Deep Learning.}
Since the breakthrough of deep learning methods in many areas of image processing, they have also been applied for flame segmentation. These methods represent the current state-of-the-art in terms of segmentation quality. In comparison to traditional machine learning methods, they no longer rely on hand-crafted features, they learn the features best suited to separate foreground and background through training. In this regard, \citet{zhong2018convolutional} is a hybrid method, since it first determines candidate regions for flames based on a threshold on color values and then uses a Convolutional Neural Network (CNN) to classify the candidate regions into flame or background. \citet{grosskopf2021evaluation} introduce a pure CNN-based method to segment industrial burner flames in images. They not only focus on binary segmentation of flames, but also perform multi-class segmentation. They also provide a public image dataset with ground truth labels. Recently, \citet{landgraf2022evaluation} analyzed several self-supervised learning methods as pre-training for the segmentation of flames. They show how the required amount of training data can be significantly reduced without decreasing the quality of the segmentation. Most recently, \citet{wang2022comparative} compared four different deep learning methods, namely Fully Convolutional Network \citep{long2015fully}, U-Net \citep{ronneberger2015u}, PSPNet \citep{zhao2017pyramid} and DeepLabV3+ \citep{chen2018encoder} on the segmentation of forest fires in images captured by UAVs. Their analysis shows that U-Net achieves the best performance, though also having the slowest inference time. Compared to that, DeepLabV3+ is slightly faster but also yields slightly worse results. In contrast, the FCN and PSPNet models have shown to be less suitable because of lower scores for the performance metrics. 


\section{Methodology}\label{Methodology}
In the following, we provide an overview of our comparative study, as well as explain the examined traditional image processing, traditional machine learning and deep learning methods in more detail. 

\subsection{Overview}
In contrast to the related work on flame segmentation and as Figure \ref{fig:overview} shows, we provide a comprehensive study that covers methods from traditional image processing to traditional machine learning and modern deep learning. Our goal is to provide helpful insights and guidance for researchers and practitioners aiming to select an appropriate approach for the binary segmentation of industrial burner flames and similar applications.  

For our study, we compare seven popular binary segmentation approaches:
\begin{enumerate}
    \item Global Thresholding (GTH) \citep{lee1990comparative},
    \item Region Growing (RG) \citep{adams1994seeded},
    \item Support Vector Machines (SVM) \citep{hearst1998support},
    \item Random Forest (RF) \citep{breiman2001random},
    \item Multilayer Perceptron (MLP) \citep{mcclelland1987parallel},
    \item U-Net \citep{ronneberger2015u},
    \item DeepLabV3+ (DL3+) \citep{chen2018EncoderDecoderAtrous}.
\end{enumerate}

The hyperparameters of the traditional image processing techniques, GTH and RG, as well as the traditional machine learning methods RF, SVM, and MLP, are tuned on the training dataset. On the other hand, all of the deep learning models are trained on the same fixed set of hyperparameters to limit the computational cost. In return, we evaluate the effect of training from scratch in comparison to fine-tuning from ImageNet pre-training and estimate the impact of the model size, through training both U-Net and DL3+ with three different backbones each.

\subsection{Traditional Image Processing}
Traditional image processing offers an easy and effective approach for binary segmentation if they are adapted for the underlying application. This is why we explore GTH and RG as a baseline for the more sophisticated data-driven methods. We implemented both GTH and RG with HALCON \citep{halcon}.

\textbf{Global Thresholding.} GTH is a widely-used method for performing binary image segmentation, where two threshold values are selected to separate the foreground objects from the background. In our case, each pixel is classified as industrial burner flame based on the following condition:
\begin{equation}
    T_{L} \le g \le T_{U}, 
\end{equation}
where $g$ is the gray value of the respective pixel between $0$ and $255$, $T_{L}$ is the lower, and $T_{U}$ is the upper threshold. To determine the optimal values for the lower and upper thresholds, we performed a grid search on the training dataset. 

\textbf{Region Growing.} RG is another traditional image processing technique for binary segmentation, which can start at any pixel in the image. It then grows the segmented region by including neighboring pixels that satisfy certain criteria. In our case, a pixel is added to the region if the following criterion is met:
\begin{equation}
    | r_{\overline{g}} - g | \le T,
\end{equation}
where $r_{\overline{g}}$ is the mean gray value of the region, $g$ is the gray value of the respective pixel, and $T$ is the chosen threshold. We segment the background with the RG and invert the segmentation results instead of segmenting the flames directly. This results in a better performance because industrial burner flames are not always coherent. For the starting pixel, we choose the lowest gray value in every given image and to determine the optimal value of the threshold, we performed a grid search on the training dataset.

\subsection{Traditional Machine Learning}
In addition to the aforementioned traditional image processing techniques, we also explored three traditional machine learning approaches: SVM, RF, and MLP. We implemented all of them in Python with Scikit-learn \citep{pedregosa2011scikit}, Scikit-image \citep{van2014scikit} and OpenCV \citep{bradski2000opencv}.

\begin{figure}
    \centering
    \includegraphics[width=\linewidth]{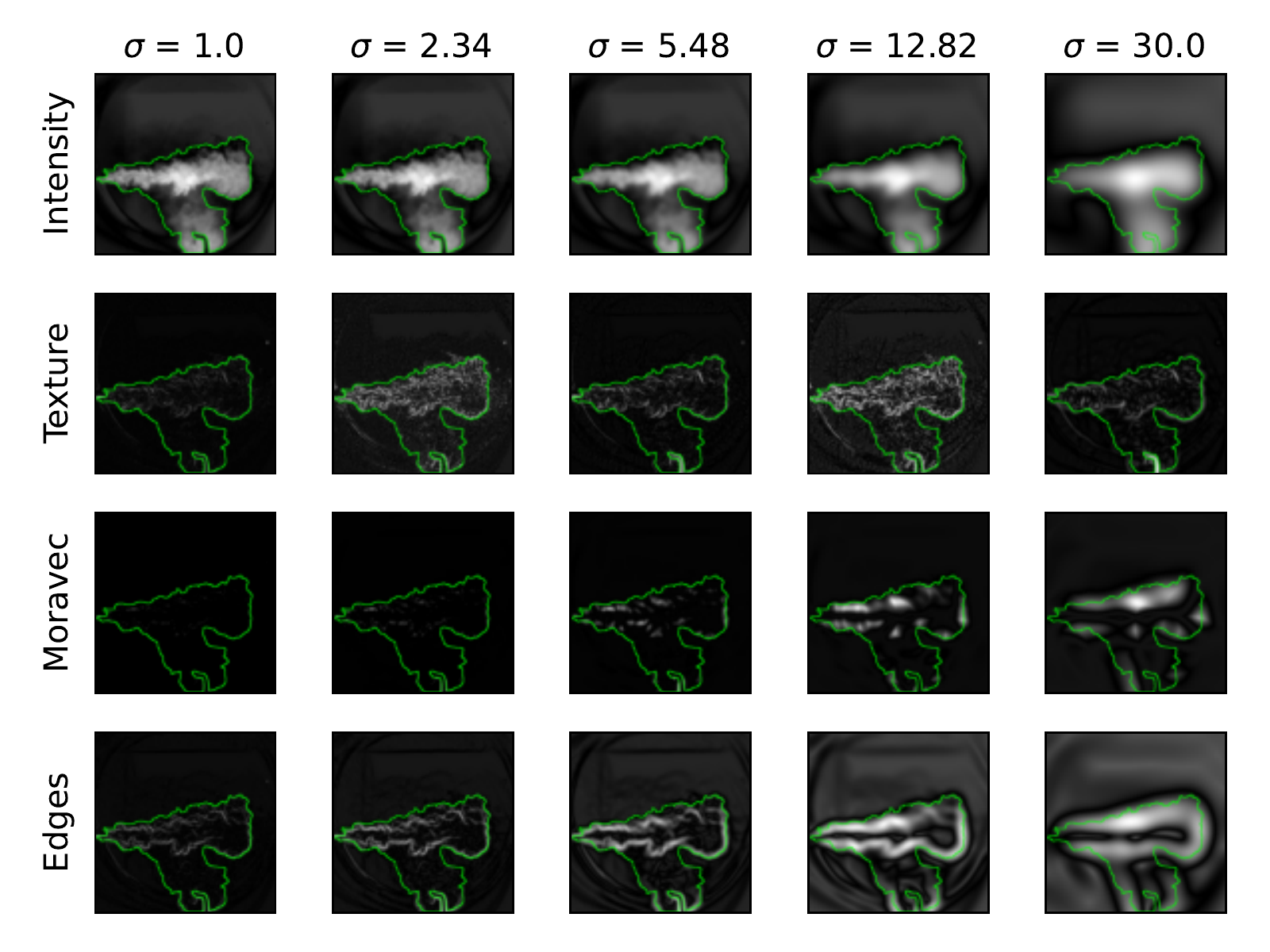}
    \caption{Example of the used basic multiscale features and Moravec corners from Scikit-image for a single image with the ground truth label's contour overlaid in green.}
    \label{fig:features}
\end{figure}

\textbf{Features.} To improve the performance of the traditional machine learning classifiers, we extract 23 hand-crafted features from the images. Like Figure \ref{fig:features} shows, these include 20 intensity, texture, edge, and corner features obtained using basic multiscale features and Moravec corners from Scikit-image. For smoothing the image with a Gaussian kernel, we used standard deviations from 1.0 to 30.0. We also use OpenCV to build three median features with square filters of sizes 51, 101, and 151. All of the features are extracted pixelwise and as a consequence, the classification is performed on each pixel individually.

\textbf{Support Vector Machines.}
SVM is a popular algorithm for binary segmentation that works by finding the optimal hyperplane that separates the positive and negative examples with the largest possible margin. For hyperparameter optimization, we performed a grid search on the training dataset to determine the optimal values for the regularization parameter and the kernel function.   

\textbf{Random Forest.}
RF is an ensemble learning method that combines multiple decision trees to improve the performance of the classifier. Thereby, each decision tree is trained on a random subset of the dataset. For hyperparameter optimization, we performed a grid search on the training dataset to determine the optimal values for the number of decision trees, the maximum depth of the trees, and the maximum number of training images for each tree.

\textbf{Multilayer Perceptron.}
MLP is a basic neural network that consists of multiple layers of interconnected nodes that can be utilized for binary segmentation. Hereby, each node applies a nonlinear activation function to the weighted sum of its inputs. Following the previous machine learning classifiers, we used the 23 extracted features as input for the MLP. To optimize the hyperparameters of the MLP, we performed a grid search on the training dataset using a range of possible values for the number of hidden layers, the number of nodes in each layer and the learning rate.

\subsection{Deep Learning}
Aside from traditional image processing techniques and traditional machine learning methods, we also explore deep learning approaches for binary segmentation of industrial burner flames. Both U-Net and DL3+ were implemented with PyTorch \citep{paszke2019PyTorchImperative}. 

\textbf{U-Net.} U-Net is a fully convolutional neural network that uses a contracting path to capture the image context and a symmetric expanding path to achieve precise localization. The resulting encoder-decoder architecture has a U-shape, which utilizes skip connections that allow information to be propagated from the encoder to the decoder. The U-Net was first introduced for medical image segmentation \citep{ronneberger2015u}. 

\textbf{DeepLabV3+.} DeepLabV3+ is a fully convolutional neural network that uses atrous (or dilated) convolutions within the atrous spatial pyramid pooling (ASPP) module to capture multi-scale contextual information. It builds upon the encoder-decoder architecture by fusing high-level ASPP features with low-level features from earlier layers in the network \citep{chen2018EncoderDecoderAtrous}.

\textbf{Backbones and Initialization.} In order to estimate the impact of the model size, we train both U-Net and DL3+ with three different backbones:
\begin{enumerate}
    \item MobileNetV3 Small (MN) \citep{howard2019searching},
    \item ResNet-18 (RN18) \citep{he2016deep},
    \item ResNet-101 (RN101) \citep{he2016deep}.
\end{enumerate}

Additionally, we evaluate the impact of the initialization for every model by:
\begin{enumerate}
    \item Training from scratch, i.e. with random weights (R),
    \item Training from ImageNet (I) \citep{deng2009ImageNetLargescale}.
\end{enumerate}

\textbf{Implementation Details.} We train all the deep learning models with a binary cross-entropy loss and employ a Stochastic Gradient Descent (SGD) optimizer based on \citet{robbins1951stochastic} with an initial learning rate of 0.01, momentum of 0.9, and weight decay of 0.0005 as optimizer-specific hyperparameters. During training, the learning rate decays based on:
\begin{equation}
lr = lr_{initial} \cdot (1 - \frac{iteration}{total\:iterations})^{0.9},
\end{equation}
where $lr$ is the current learning rate, and $lr_{initial}$ is the initial learning rate. All models are trained for 25 epochs with a batch size of 8 and without any data augmentations to ensure a fair comparison.

\section{Experiments}
\begin{figure}[t!]
    \centering
    \subfigure{\includegraphics[width=0.32\linewidth]{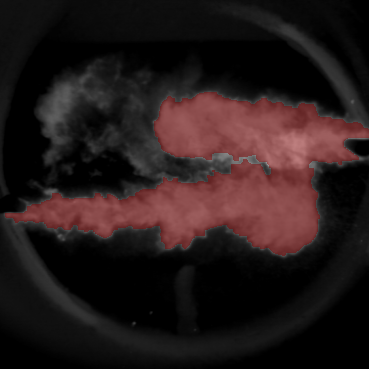}}
    \subfigure{\includegraphics[width=0.32\linewidth]{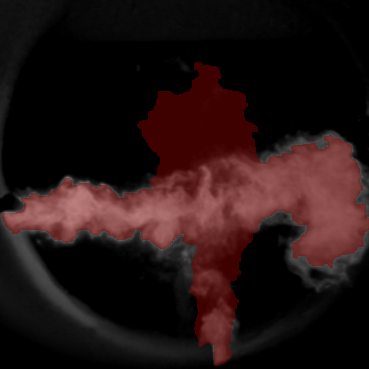}}
    \subfigure{\includegraphics[width=0.32\linewidth]{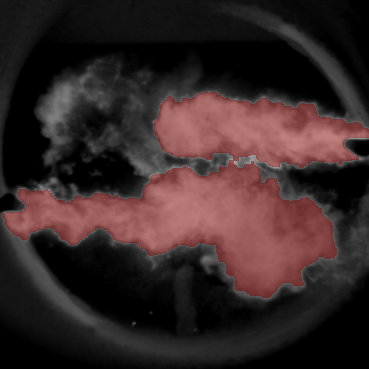}}
    \subfigure{\includegraphics[width=0.32\linewidth]{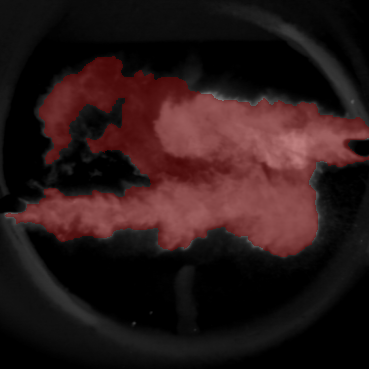}}
    \subfigure{\includegraphics[width=0.32\linewidth]{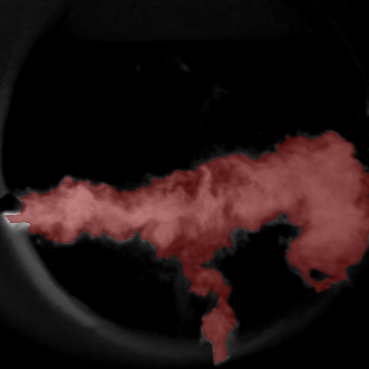}}
    \subfigure{\includegraphics[width=0.32\linewidth]{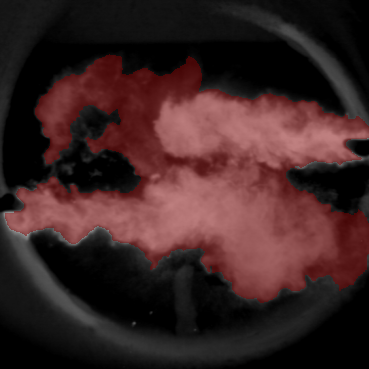}}
    \caption{Example images of the industrial burner flames dataset with the ground truth labels overlaid as transparent red regions. The upper row displays the original labels by \citet{grosskopf2021evaluation}, whereas the lower three images represent the labels that we created.}
    \label{dataset_comparison}
\end{figure}

\begin{figure*}[h!]
    \centering
    \subfigure[Input Image]{\includegraphics[width=0.24\linewidth]{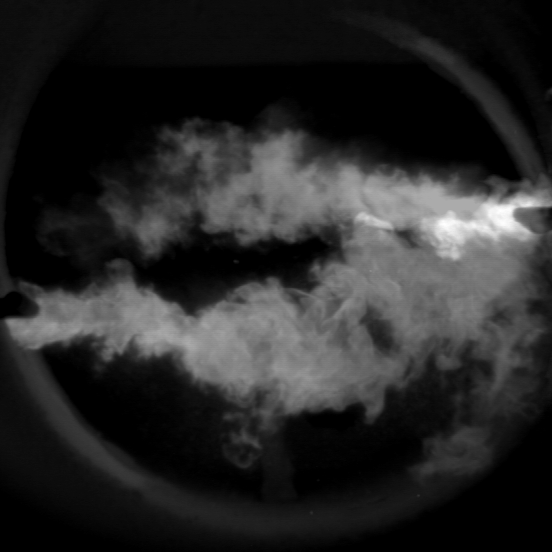}}
    \subfigure[Ground Truth]{\includegraphics[width=0.24\linewidth]{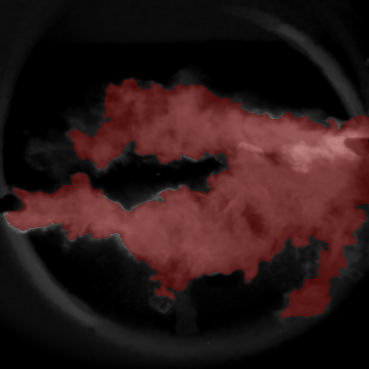}}
    \subfigure[GTH]{\includegraphics[width=0.24\linewidth]{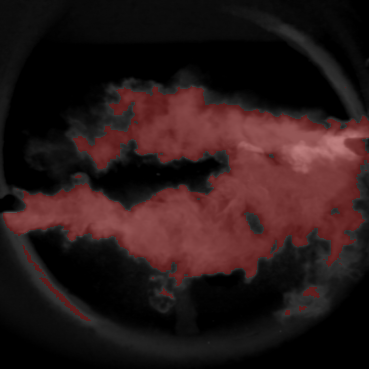}}
    \subfigure[RG]{\includegraphics[width=0.24\linewidth]{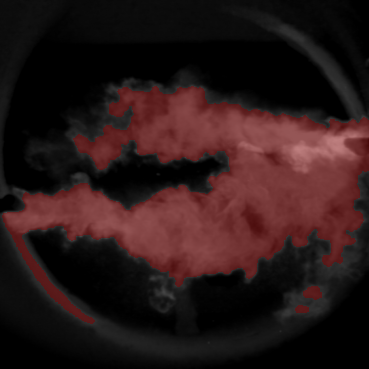}}
    \subfigure[SVM]{\includegraphics[width=0.24\linewidth]{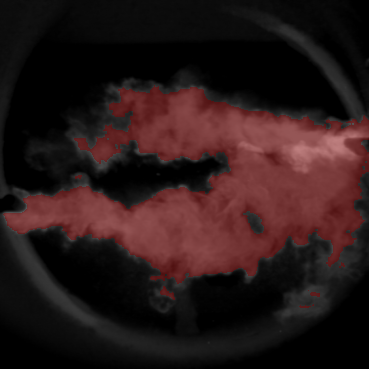}}
    \subfigure[RF]{\includegraphics[width=0.24\linewidth]{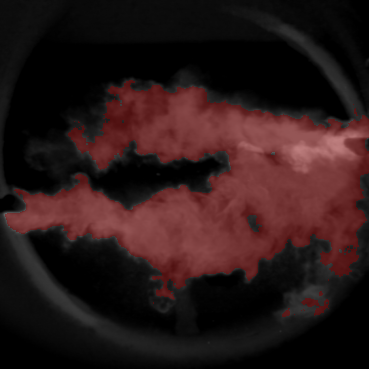}}
    \subfigure[MLP]{\includegraphics[width=0.24\linewidth]{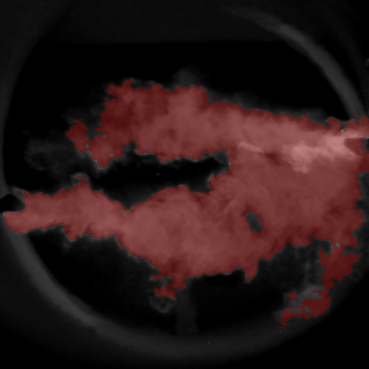}}
    \subfigure[DL3+ (RN18-I)]{\includegraphics[width=0.24\linewidth]{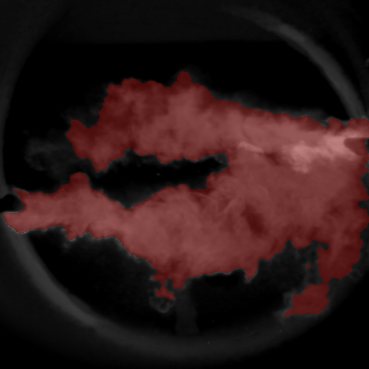}}
    \caption{Qualitative comparison between the input image (a) overlaid in transparent red color with the ground truth (b), the segmentation results of Global Thresholding (c), Region Growing (d), Support Vector Machines (e), Random Forest (f), Multilayer Perceptron (g), and DeepLabV3+ with a ResNet-18 backbone pre-trained on ImageNet (h).}
    \label{qualitative_comparison}
\end{figure*}

In this section, we share a variety of experiments conducted on the basis of a modified public dataset to gather helpful insights about the strengths and limitations of all the methods in our study. 

\subsection{Dataset}
All of our experiments are based on a modified version of the publicly available industrial burner flames dataset provided by \citet{grosskopf2021evaluation}. The original dataset consists of 3000 labeled grayscale images of two industial burner flames with 552 $\times$ 552 pixels in size.

In Figure \ref{dataset_comparison}, the upper row shows that some labels in the original dataset are questionable. We suspect that these labels were at least in part created automatically. To address this, we randomly selected 200 images and relabeled them by hand, creating a new dataset of 160 training images and 40 test images. As shown in the lower row of Figure \ref{dataset_comparison}, this process resulted in an improved label quality. In the original labels, flames represented 23.8\% of the dataset, whereas in our labels the portion of flames is 26.5\%. Finally, we computed the Intersection over Union (IoU) between the original labels and our labels, which yielded a score of 80.8\%.

\subsection{Quantitative Evaluation}
\setlength{\extrarowheight}{3pt}
\begin{table}
\begin{center}
\begin{tabular}{lrr}
& IoU [\%] $\uparrow$ & Inference Time [ms] $\downarrow$\\ \hline 
GTH & 80.3 & 0.1 \\
RG & 77.0 & 13.4 \\ \hdashline
SVM & 82.3 & 8.2 \\
RF & 87.0 & 124.8 \\
MLP & 86.6 & 106.3 \\ \hdashline
U-Net (MN) & 92.6 (91.9) & 106.7 (7.6) \\
U-Net (RN18) & 92.9 (92.6) & 138.4 (5.6) \\
U-Net (RN101) & 93.1 (91.9) & 436.9 (16.6) \\
DL3+ (MN) & 92.6 (91.9) & 81.2 (6.5) \\
DL3+ (RN18) & 93.2 (92.9) & 117.8 (4.6) \\
DL3+ (RN101) & 92.9 (92.3) & 394.3 (15.5) \\
\end{tabular}
\end{center}
\caption{Quantitative comparison between the traditional image processing, traditional machine learning, and deep learning Intersection over Union (IoU) and inference time per image. IoU scores in parantheses refer to training from scratch, whereas the regular scores depict fine-tuning from ImageNet. Inference times in parantheses were computed on a GPU, while the regular values are CPU times.}
\label{quantitative_comparison}
\end{table}

Table \ref{quantitative_comparison} presents a quantitative comparison between the best segmentation results obtained by each method in our study, along with their respective inference time per image. Unless otherwise noted, we used an AMD EPYC 7502 32-Core CPU with 1 TB of RAM. The results demonstrate that there are clear improvements going from traditional image processing (GTH: 80.3\%) over traditional machine learning (RF: 87.0\%) to deep learning (DL3+ (RN18-I): 93.2\%). There is, however, a corresponding increase in CPU time going from 0.1 ms per image with GTH up to 436.9 ms with U-Net (RN101-I). 

\textbf{Traditional Image Processing.} Looking at the traditional image processing techniques in particular, GTH outperforms RG in both the segmentation task and the needed inference time. GTH yields an IoU score of 80.3\% which is 3.3\% higher than what RG achieves. On top of that, GTH only needs 0.1 ms to process an image, whereas RG needs 13.4 ms.  

\textbf{Traditional Machine Learning.} In comparison to the traditional image processing techniques, the machine learning approaches, SVM, RF, and MLP, yield a significant performance improvement in the segmentation task. SVM achieved an IoU score of 82.3\%, outperforming both GTH and RF by noteworthy margin. In addition, SVM only requires 8.2 ms to process an image, making it faster than RF but slower than GTH. With 87.0\% RF achieved the highest IoU score among the traditional machine learning methods. However, it also requires 124.8 ms to process an image. MLP achieved a very similar IoU score of 86.6\%, but with a slightly faster inference time of 106.3 ms.   

\textbf{Deep Learning.} 
Judging by the results laid out in Table \ref{quantitative_comparison}, all of the deep learning approaches outperform the traditional image processing and machine learning methods on the segmentation task. Depending on the configuration, the models achieve IoU scores between 91.9\% and 93.2\%. Table \ref{quantitative_comparison} also shows that training from scratch in comparison to fine-tuning from ImageNet pre-training deteriorates performance by 0.3\% to 1.2\%. As Table \ref{quantitative_comparison} displays, deep learning comes at a high computational cost. Inference times on the CPU range from 81.2 ms for DL3+ (MN) to a maximum of 436.9 ms for U-Net (RN101). However, inference times can be decreased greatly if inference is performed on a GPU. In our case, using a common NVIDIA GeForce RTX 3090 GPU with 24 GB of memory reduced inference times to a range from 4.6 to 16.6 ms depending on the architecture. Among all of the deep learning models, DL3+ (RN18-I) achieved the highest IoU score of 93.2\% while simultaneously having the fastest inference time of 4.6 ms with the GPU.

\subsection{Qualitative Evaluation}
Figure \ref{qualitative_comparison} shows a visual comparison between the traditional image processing techniques, the traditional machine learning approaches, and the best deep learning model. Overall, the qualitative evaluation of the segmentation results confirms the observations of the quantitative analysis: There are clear improvements going from traditional image processing to traditional machine learning to deep learning. Figure \ref{more_examples} in the Appendix shows more qualitative examples, which corroborate this claim. 

\textbf{Traditional Image Processing.} The two traditional image processing techniques, GTH and RG, largely suffer from the same shortcoming: Both oversegment the industrial burner chamber and undersegment the flame. As both methods solely rely on the underlying gray values, there is no way for these methods to distinguish between bright areas of the image that belong to the chamber or darker parts of the flame. Hence, the segmentation results are systematically flawed in these situations. 

\textbf{Traditional Machine Learning.} In comparison to the traditional image processing techniques, the machine learning approaches, SVM, RF, and MLP, do not suffer from oversegmentation of the chamber. However, they undersegment darker parts of the flames too, which is especially noticeable for SVM. Overall, all of the traditional machine learning methods can improve upon the traditional image processing techniques due to the extra information added by the hand-crafted features.

\textbf{Deep Learning.} Visually, an even better segmentation result is achieved by the best deep learning model DL3+ (RN18-I). The segmentation result suffers from no apparent systematic shortcomings. Through the training process, the deep learning model learns meaningful features and patterns that seem to generalize very well to the test data. This also holds for the other deep learning models not shown in Figure \ref{qualitative_comparison}.

\subsection{Impact of Training Dataset Size}
\setlength{\extrarowheight}{3pt}
\begin{table}[t!]
\begin{center}
\begin{tabular}{lrr}
& IoU [\%] $\uparrow$ & Performance Loss [\%] $\downarrow$ \\ \hline 
GTH & 78.6 & 1.7 \\
RG & 74.3 & 2.7\\ \hdashline
SVM & 80.6 & 1.7 \\
RF & 84.7 & 2.3 \\
MLP & 84.9 & 1.7 \\ \hdashline
U-Net (MN) & 91.7 (90.5) & 0.9 (1.4) \\
U-Net (RN18) & 91.6 (91.1) & 1.3 (1.5) \\
U-Net (RN101) & 92.1 (88.1) & 1.0 (3.8) \\
DL3+ (MN) & 91.4 (90.6) & 1.2 (1.3) \\
DL3+ (RN18) & 91.8 (91.6) & 1.4 (1.3) \\
DL3+ (RN101) & 91.8 (90.7) & 1.1 (1.6) \\
\end{tabular}
\end{center}
\caption{Quantitative comparison between the traditional image processing, traditional machine learning, and deep learning IoU with just 40 training images and 160 test images. Additionally, the performance loss compared to the regular dataset with 160 training images is displayed. The IoU scores in parantheses refer to training from scratch, whereas the regular scores depict fine-tuning from ImageNet.}
\label{quantitative_comparison_dataset_a}
\end{table}

In order to evaluate the effect of the number of training images, we retuned all of the traditional image processing and traditional machine learning methods and retrained all deep learning models with an inverse ratio of training and test images, i.e. we use just 40 randomly selected images for training and the remaining 160 images for testing.

As Table \ref{quantitative_comparison_dataset_a} shows, the observations from the previous experiments remain unchanged: There are significant segmentation improvements going from traditional image processing (GTH: 78.6\%) to traditional machine learning (MLP: 84.9\%) to deep learning (U-Net (RN101-I): 92.1\%). To our surprise, however, the average performance loss was larger for the traditional image processing techniques and traditional machine learning approaches.

GTH and RG suffered a performance loss of 1.7\% and 2.7\%, respectively. The IoU of the traditional machine learning-based methods deteriorated by 1.9\% on average. Whereas the average performance loss for the deep learning models pre-trained on ImageNet was only 1.2\% and 1.8\% for the models trained from scratch.

\section{Discussion}
We conducted extensive experiments to compare traditional image processing techniques with traditional machine learning and deep learning methods. In the following, we discuss our key insights into the strengths and limitations of each approach. 

\textbf{Traditional Image Processing.} GTH and RG are both effective for creating a baseline binary segmentation of industrial burner flames. They are computationally efficient, require little tuning and offer fast inference times, making them a good choice when speed is a priority. In our experiments, GTH outperformed RG on the segmentation task, with the fastest inference time of all examined methods with just 0.1 ms per image. However, these techniques may not be suitable if the quality of the segmentation results is more important, as they rely solely on gray values and can suffer from under- and oversegmentation.  

\textbf{Traditional Machine Learning.} SVM, RF, and MLP significantly improve segmentation results compared to GTH by using hand-crafted features. For example, RF achieved a 6.7\% relative IoU score improvement over GTH. However, these methods are computationally more expensive and require careful engineering and feature selection. It is worth noting that adding more features could improve the performance. On the other hand, it can also introduce noise and lead to overfitting, aside from increasing computation time. As a possible solution, SVM and RF allow the computation of feature importances to help select appropriate features.

\textbf{Deep Learning.} Our experiments show that the deep learning models U-Net and DL3+ achieve the best segmentation performance overall with a maximum IoU score of 93.2\%. Surprisingly, the choice of architecture or pre-training on ImageNet had very little impact on the performance, although the latter helped in all experiments. Another revealing advantage of deep learning is its ability to generalize from small datasets, as the models suffered the smallest performance loss when trained on less data. The only apparent downside to the deep learning methods is their high computational cost. However, training and inference times can greatly benefit from using a GPU. Overall, deep learning is the best choice for binary segmentation of industrial burner flames, particularly if a GPU is available.

\section{Conclusion}
In this work, we conducted a comparative study of traditional image processing techniques, traditional machine learning methods, and deep learning approaches for the binary segmentation of industrial burner flames. Extensive experimentation on a modified version of a public dataset \citep{grosskopf2021evaluation} revealed key insights into the strengths and limitations of each approach. 

Traditional image processing techniques like Global Thresholding and Region Growing offer a fast and simple solution but suffer from systematic under- and oversegmentation. Traditional Machine learning methods, such as Support Vector Machines, Random Forests, and Multilayer Perceptrons, improve the segmentation performance in exchange for increased computational cost and feature engineering effort. Deep learning models, like U-Net and DeepLabV3+, achieve the best segmentation performance in our study while showing remarkable ability to generalize even from small datasets. Despite their high computational requirements, they are the best choice for binary segmentation of industrial burner flames, especially if a GPU is available.

In summary, our study provides helpful insights and guidance for researchers and practitioners aiming to select an appropriate approach for the binary segmentation of industrial burner flames and beyond. For the highest accuracy, deep learning is the leading approach, while for fast and simple solutions, traditional image processing techniques remain a viable option. With the continuing progress in deep learning and the increasing availability of compute power, we expect these methods to become even more capable and efficient in the future.

\section*{Acknowledgment}
This research is funded by Invest BW on behalf of the Ministry of Economics, Labor and Tourism of Baden-Württemberg.

{
	\begin{spacing}{1.17}
		\normalsize
		\bibliography{ISPRSguidelines_authors_fullpaper} 
	\end{spacing}
}

\onecolumn
\section*{APPENDIX}
\label{sec:appendix}
\begin{figure}[ht!]
\centering
    \subfigure{\includegraphics[width=0.24\linewidth]{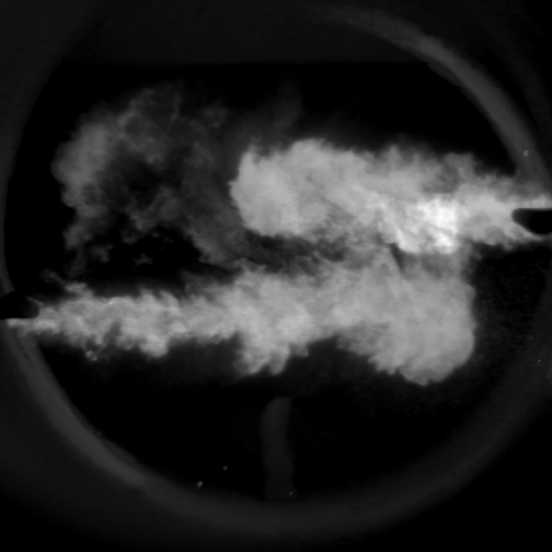}}
    \subfigure{\includegraphics[width=0.24\linewidth]{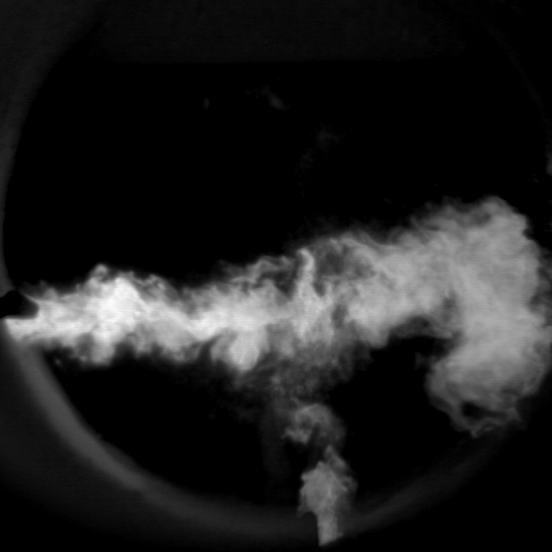}}
    \subfigure{\includegraphics[width=0.24\linewidth]{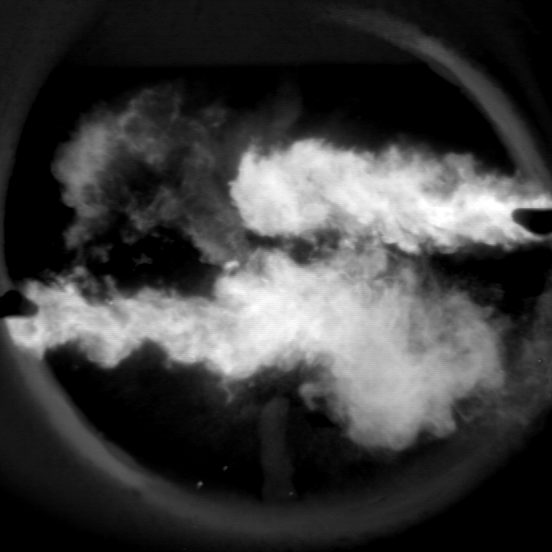}}
    \subfigure{\includegraphics[width=0.24\linewidth]{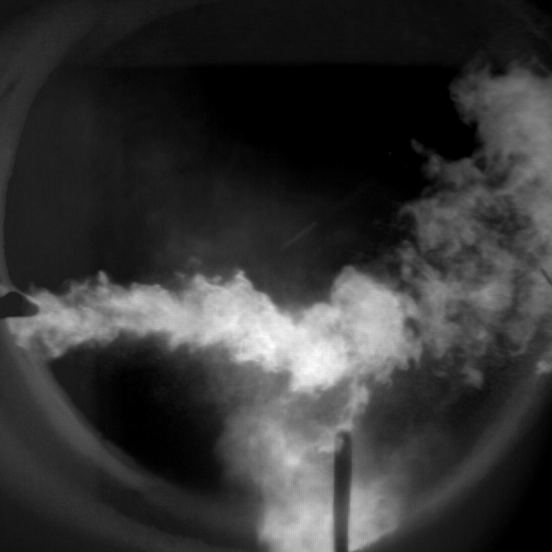}}
    \subfigure{\includegraphics[width=0.24\linewidth]{figures/qualitative_examples/ground_truth_000091.png}}
    \subfigure{\includegraphics[width=0.24\linewidth]{figures/qualitative_examples/ground_truth_000119.png}}
    \subfigure{\includegraphics[width=0.24\linewidth]{figures/qualitative_examples/ground_truth_000191.png}}
    \subfigure{\includegraphics[width=0.24\linewidth]{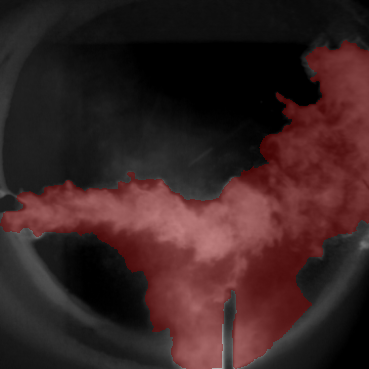}}
    \subfigure{\includegraphics[width=0.24\linewidth]{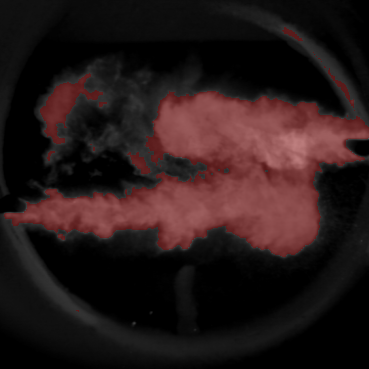}}
    \subfigure{\includegraphics[width=0.24\linewidth]{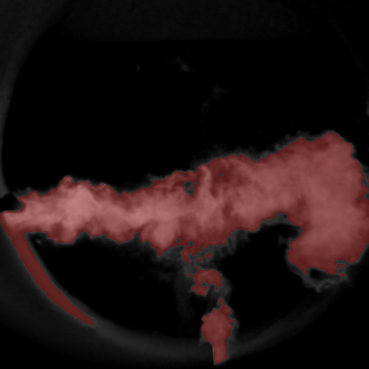}}
    \subfigure{\includegraphics[width=0.24\linewidth]{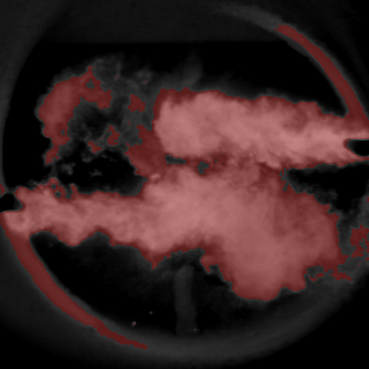}}
    \subfigure{\includegraphics[width=0.24\linewidth]{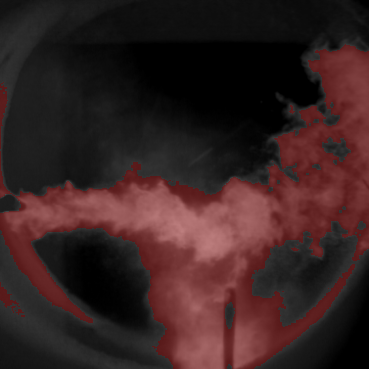}}
    \subfigure{\includegraphics[width=0.24\linewidth]{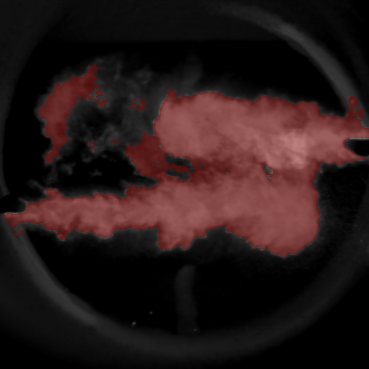}}
    \subfigure{\includegraphics[width=0.24\linewidth]{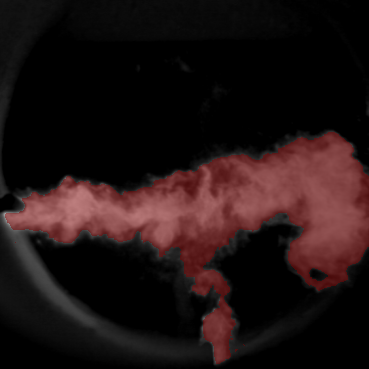}}
    \subfigure{\includegraphics[width=0.24\linewidth]{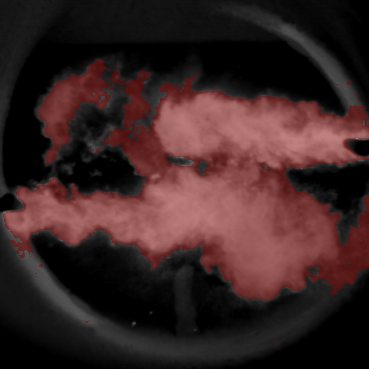}}
    \subfigure{\includegraphics[width=0.24\linewidth]{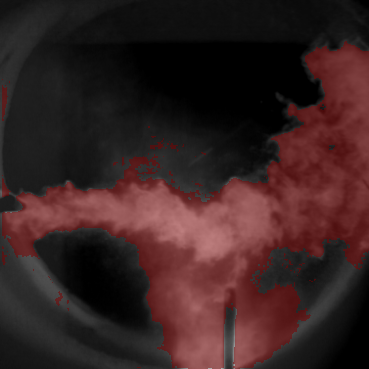}}
    \subfigure{\includegraphics[width=0.24\linewidth]{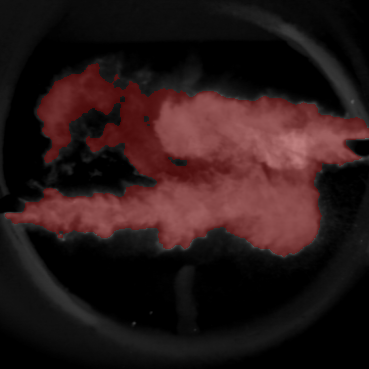}}
    \subfigure{\includegraphics[width=0.24\linewidth]{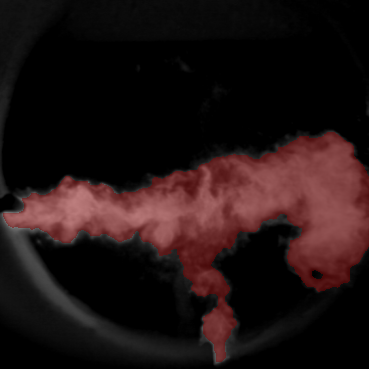}}
    \subfigure{\includegraphics[width=0.24\linewidth]{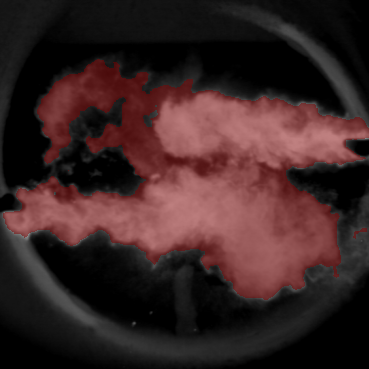}}
    \subfigure{\includegraphics[width=0.24\linewidth]{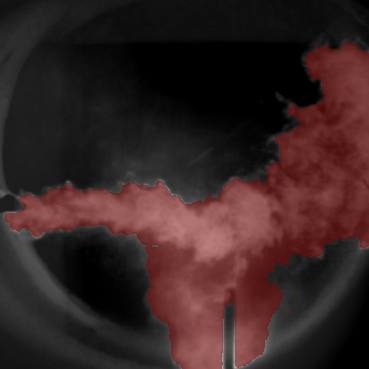}}
\caption{More qualitative comparisons between the input image in the first row, the ground truth in the second row, the segmentation results of GTH in the third row, RF in the fourth row, and DL3+ (RN18-I) in the last row.}
\label{more_examples}
\end{figure}

\end{document}